\documentclass[10pt,twocolumn,letterpaper]{article} 

\usepackage{avss}
\usepackage{times}
\usepackage{epsfig}
\usepackage{graphicx}
\usepackage{amsmath}
\usepackage{amssymb}
\usepackage{multirow}
\usepackage{array}

\usepackage[pagebackref=true,breaklinks=true,letterpaper=true,colorlinks,bookmarks=false]{hyperref}

\avssfinalcopy 

\def\avssPaperID{29} 

\ifavssfinal\fi
\begin{document}
\newcolumntype{C}[1]{>{\centering\arraybackslash}m{#1}}
\title{CPNet: Cross-Parallel Network for Efficient Anomaly Detection}

\author{Youngsaeng Jin\\
Korea University\\
Seoul, Korea\\
{\tt\small Youngsjin@korea.ac.kr}

\and
Jonghwan Hong\\
Korea University\\
Seoul, Korea\\
{\tt\small jhong2661@korea.ac.kr}

\and
David Han\\
Drexel University\\
Philadelphia, PA USA\\
{\tt\small dkh42@drexel.edu}

\and
Hanseok Ko\\
Korea University\\
Seoul, Korea\\
{\tt\small hsko@korea.ac.kr}
}

\maketitle

\begin{abstract}
   Anomaly detection in video streams is a challenging problem because of the scarcity of abnormal events and the difficulty of accurately annotating them. To alleviate these issues, unsupervised learning-based prediction methods have been previously applied. 
   These approaches train the model with only normal events and predict a future frame from a sequence of preceding frames by use of encoder-decoder architectures so that they result in small prediction errors on normal events but large errors on abnormal events. The architecture, however, comes with the computational burden as some anomaly detection tasks require low computational cost without sacrificing performance. 
   In this paper, Cross-Parallel Network (CPNet) for efficient anomaly detection is proposed here to minimize computations without performance drops. It consists of \(N\) smaller parallel U-Net, each of which is designed to handle a single input frame, to make the calculations significantly more efficient. Additionally, an inter-network shift module is incorporated to capture temporal relationships among sequential frames to enable more accurate future predictions. The quantitative results show that our model requires less computational cost than the baseline U-Net while delivering equivalent performance in anomaly detection.
\end{abstract}

   
\section{Introduction}

Anomaly detection in videos is the identification of unexpected or unfamiliar events from video streams containing mostly normal or expected activities. With the great success of machine learning methodologies in other applications, automated detection of abnormalities in videos has attracted much interest in computer vision communities. It is a challenging task for supervised learning as manually annotating frame-wise labels for normality/abnormality and spatially localizing them are labor-intensive. Additionally, collecting abnormal events in the dataset is difficult, as by definition abnormal events are rare in real life. Thus, it leads to a class imbalance in the training set between normal and abnormal events. 

For these reasons, unsupervised learning~\cite{eskin2002geometric, ahmad2017unsupervised} is the method of choice in anomaly detection. These approaches first train the models only with normal events. The expectation is that when an abnormal event is shown to the model, its output would be quite different from the model's response to normal events. Most of these models employ a network composed of an encoder and a decoder such that the network trained with only normal events would reconstruct the input image with minimal error~\cite{hasan2016learning, zhao2017spatio,luo2017revisit}. As the model is trained by normal events only, the video frames with small reconstruction errors are considered as normal events while those with large reconstruction errors are classified as abnormal events. Thus, the reconstruction error is used as an indicator of abnormal events. The key to this approach is to ensure that the model overfits to the normal events such that it would perform poorly in reconstructing inputs of abnormal events.

To weaken the model's generalizability, a future prediction approach~\cite{liu2018future} has been proposed. Given a video clip, instead of reconstructing a current frame, this approach predicts a future frame based on its several preceding frames. As the input frames do not contain a target(future) frame to be predicted, the task is harder compared to the task of input image reconstruction. By making the task harder, the approach leads to much larger prediction errors on abnormal events compared to the error in the case of normal events used in the training.

U-Net~\cite{unet} architecture, which exhibits good performance on image-to-image tasks such as segmentation and super-resolution, has been widely used in previous anomaly detection methods~\cite{liu2018future,nguyen2020anomaly,park2020learning,tang2020integrating}. When feeding multiple frames as an input in the prediction approach, these frames are concatenated through the channel dimension. As such, the network can capture both spatial and temporal correlations so that future frames can be predicted more accurately. The architecture, however, comes with the computational burden of having a large number of input channels as each input is composed of multiple frames stacked together over the channel dimension. 

To alleviate this burden, we propose Cross-Parallel Network (CPNet) which is composed of \(N\) parallel U-Net (U-PARL) with each U-Net designed to handle only a single temporal frame. The number of feature maps (channel dimensions) of both the input and the output of the convolution process is reduced by \(N\), resulting in a computational reduction of \(\frac{1}{N^2}\) for a single path. Counting that the network is composed of \(N\) parallel networks, the overall computational cost is reduced by \(\frac{1}{N}\) compared to the conventional U-Net.
As these encoder-decoder networks process each input frame independently, capturing temporal relationships among the frames is handled only at the final stage of image reconstruction. Therefore, the architecture by design makes it more difficult to predict the future frame compared to the network that takes the stacked input frames.


This suppression of the temporal relationships among the input frames resulted in large prediction errors even for normal events. In order to extract the temporal cross-frame correlations without much increase in computational cost, we introduce a cross-network shift frame strategy~\cite{lin2019tsm}. In this approach, a part of feature maps from one input frame is shifted to those of its sequentially neighboring frames at each network layer in parallel. Although this shift operation incurs some data movement costs, it will not increase the FLOP of the overall architecture.

The contributions of our method are summarized as follows:
\begin{quote}
    1) A novel architecture CPNet for efficient and effective unsupervised anomaly detection framework is composed of a parallel U-Net architecture with inter-frame feature shifting modules.
    
    2) We introduce a novel architecture composed of smaller parallel networks, each designed to process an individual image frame. The reduction of the channel dimensions of each path resulted in a significant reduction of computational cost. 
    
    3) We introduce a cross-network feature correlation strategy by shifting a part of the feature map from one input frame to its neighbors. The method allowed an effective abstraction of cross-network temporal correlations without much increase in computational cost.
    
    4) The CPNet results in a significant reduction in computational cost while delivering equivalent performance on Ped2~\cite{sabokrou2017deep} and Avenue~\cite{lu2013abnormal} datasets.
    
\end{quote}

\section{Related work}
\paragraph{Hand-crafted Unsupervised Anomaly Detection.} In the early years of anomaly detection, hand-crafted features were used as a discriminative basis for detecting abnormal events. Some efforts utilized low-level spatial-temporal features, such as histogram of oriented gradients (HOG)~\cite{dalal2005histograms} or histogram of oriented flows (HOF)~\cite{dalal2006human}. Other methods~\cite{wu2010chaotic, tung2011goal} exploited low-level trajectory features to represent the patterns. Markov Random Field (MRF)~\cite{zhang2005semi, kim2009observe}, Mixture of Dynamic Textures (MDT)~\cite{mahadevan2010anomaly}, and Gaussian regression~\cite{cheng2015video} were also utilized to characterize the statistical distribution in obtaining decision boundaries. However, these methods are susceptible to failure in crowded or complex scenes.

\paragraph{Reconstruction Approach for Anomaly Detection.} To combat the issue of data set class imbalance, some efforts explored unsupervised approaches using deep learning architecture. The idea is by first training a model with normal events only to the point of overfitting, the model will respond quite differently when it observes abnormal input. These models typically use autoencoder-type architectures to reconstruct input images, and the reconstruction error is used as the measure to determine if an event is abnormal. 
Many reconstructive approaches based on CNN have been proposed in recent years including convolutional auto-encoder~\cite{hasan2016learning, zhao2017spatio}. As Recurrent Neural Network (RNN) is capable of temporal modeling, some methods~\cite{luo2017revisit,luo2017remembering} proposed convolutional LSTM to take advantages of CNN and RNN. Ravanbakhsh~\textit{et al.}~\cite{ravanbakhsh2017abnormal} utilized Generative Adversarial Network (GAN) to boost anomaly detection performance.
Some methods~\cite{gong2019memorizing, park2020learning} exploited memory networks to read and write normal patterns and made reconstructions biased to normal events. Zaheer~\textit{et al.}~\cite{zaheer2020old} employed an adversarial model for identifying real and fake data to distinguish good and bad quality reconstructions.
Although these methods brought significant improvements in anomaly detection, their performance often plateaued when reconstruction errors become small on abnormal events. This is primarily due to generalizing capacity of deep neural networks.

\begin{figure*}[t]
\begin{center}
    \centering 
    \includegraphics[width=0.95\textwidth]{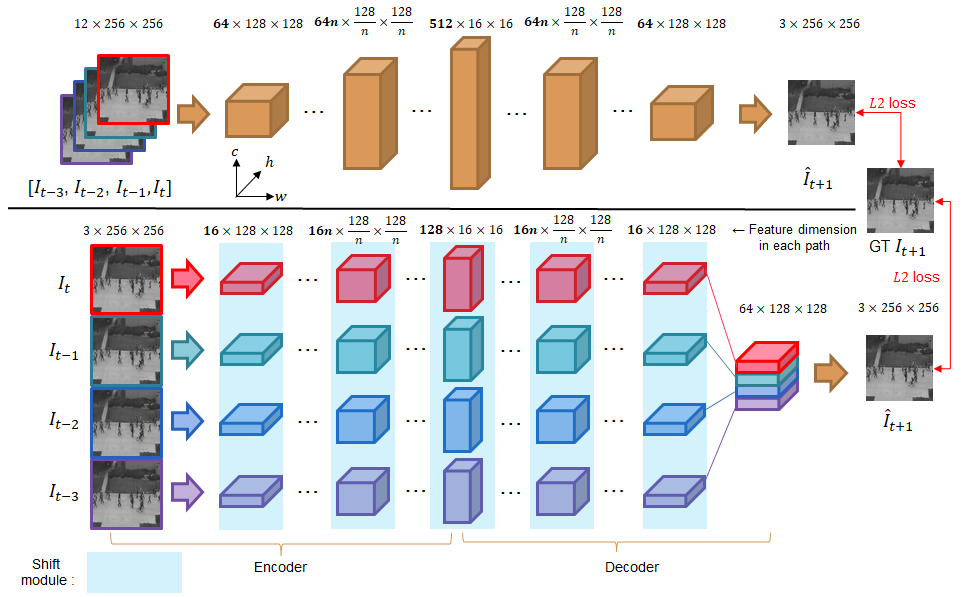}
\end{center}
\centering\caption{Illustration of the proposed method. Each cuboid denotes the feature maps from the conventional U-Net (Top) and CPNet (Bottom), which is modified U-Net. CPNet consists of four smaller (horizontally) parallel U-Nets, which process input frame independently, and (vertically) parallel shift modules, which capture temporal relationships in each layer.}
\label{fig:model}
\end{figure*}

\paragraph{Prediction Approach for Anomaly Detection.} The small error problem of deep neural networks can be addressed by making the task more challenging. Instead of the reconstruction task, a prediction task can be imposed here to force the network to predict future frames from preceding frames only. Since the input doesn't include future frames, the task is harder than the reconstruction problem and also harder to generalize with small categorical data in the training set. In unsupervised video generation tasks, Mathieu~\textit{et al.}~\cite{mathieu2015deep} employed adversarial learning to boost the performance to produce more natural frames in videos while Lotter~\textit{et al.}~\cite{lotter2016deep} predicted future frames in a video sequence with each layer in the network making local predictions and only forwarding deviations from those predictions to subsequent network layers. Chen~\textit{et al.} ~\cite{chen2017video} predicted imaginary videos by transformation generation. Motivated by these methods, the unsupervised video generation techniques were employed in unsupervised anomaly detection. Liu~\textit{et al.}~\cite{liu2018future} and Nguyen~\textit{et al.}~\cite{nguyen2020anomaly} utilized adversarial learning and added constraints in terms of appearance (intensity loss and gradient loss) and motion (optical flow loss).
Tang~\textit{et al.}~\cite{tang2020integrating} employed both reconstruction and prediction approaches for improvement.

Although these methods achieved improved performance, most of them require heavy computations as they employ conventional U-Net for reconstruction or prediction~\cite{liu2018future,nguyen2020anomaly,park2020learning,tang2020integrating}. We propose a novel modified U-Net structure with temporal shifting to reduce computational cost while delivering equivalent performance on anomaly detection.

\section{Method}

U-Net~\cite{unet} is a convolutional neural network that is developed for bio-medical image segmentation. To alleviate the problem of information loss and gradient vanishing, U-Net is proposed to add shortcuts between lower-level layers and higher-level layers with the same resolution. Due to its effectiveness, U-Net is employed in many image-to-image tasks including image prediction.
The proposed CPNet, in which modified U-Net is used as a future frame predictor for efficient anomaly detection, is illustrated in Figure~\ref{fig:model}. In the following subsections, we will describe our approach in detail.

\subsection{Future Frame Predictor}
Given a video with \(t\) consecutive frames \(I_1, I_2,...,I_t\in R^{3\times W \times H}\), the predictor predicts a future frame \(I_{t+1}\). For training the network, \(L_2\) distance between the predicted frame \(\hat{I}_{t+1}\) and the ground truth \(I_{t+1}\) is applied.

\begin{equation} \label{eq:2}
L = ||\hat{I}_{t+1}-I_{t+1}||^2_2.
\end{equation}
Many of the existing prediction based anomaly detection methods~\cite{liu2018future,nguyen2020anomaly,tang2020integrating} employ four consecutive frames \([I_{t-3}, I_{t-2},I_{t-1},I_{t}]\in R^{12\times W \times H}\) stacked together as the input to the prediction network. Convolutional operation on a stack of sequential frames strongly captures temporal relationships among them, thus it can predict future frame \(I_{t+1}\) accurately. In order to make the prediction task harder, our network processes each input frame independently by four identical but independent network pathways and concatenates them at the final layer of image prediction as shown in Figure~\ref{fig:model}. The burden of extracting temporal relationships among the frames fall on the final convolution layer, thus its prediction capability is severely limited by design.

By dividing the U-Net into four smaller parallel networks and having each parallel network process a single input frame, the size of the output feature maps(channel dimension) is \(\frac{1}{4}\) compared to the ones produced when all four frames are stacked as an input.


For a convolutional layer \(F_{out}=Conv(F_{in})\) with 
an input feature map \(F_{in}\in R^{C_{in}\times W\times H}\) and an output feature map \(F_{out}\in R^{C_{out}\times W\times H}\), its computational cost is

\begin{equation} \label{eq:3}
O=C_{in}*C_{out}*K^2*W*H.
\end{equation}
In our network, as each smaller U-Net has \(\frac{1}{4}\) channel maps in comparison with the existing methods, e.g., \(F_{in}\in R^{C_{in}\times W\times H\longrightarrow \frac{C_{in}}{4}\times W\times H}\) and \(F_{out}\in R^{C_{out}\times W\times H\longrightarrow \frac{C_{out}}{4}\times W\times H}\), its computational cost is reduced as

\begin{equation} \label{eq:4}
O'=\frac{C_{in}}{4}*\frac{C_{out}}{4}*K^2*W*H=\frac{O}{16}.
\end{equation}

As there are four smaller U-Nets in our predictor, its net computational cost is \(\frac{1}{16}\times 4=\frac{1}{4}\) of the existing prediction methods, which employed the conventional U-Net as a predictor, theoretically. 


\subsection{Inter-frame Feature Map Shift}
\begin{figure}[t]
\begin{center}
    \centering 
    \includegraphics[width=0.30\textwidth]{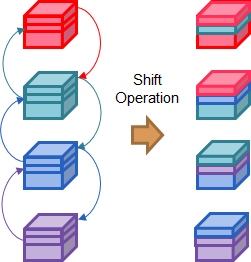}
\end{center}
\centering\caption{Illustration of the shift operation. This operation shifts feature maps from a frame to those of its neighboring frames.}
\label{fig:shift}
\end{figure}

Relying on the last convolutional layer for extracting temporal relationships among the frames turned out to be too challenging even for normal events. To alleviate this issue, we employ an inter-network shift module~\cite{lin2019tsm}, which shifts parts of the feature maps to neighboring frames to capture temporal relationships among consecutive frames. The operation of the shift module is illustrated in Figure~\ref{fig:shift}. As in ~\cite{lin2019tsm}, \(\frac{1}{4}\) feature maps from a frame-shift to those of its neighboring frames (\(\frac{1}{8}\) for each direction). This shift operation is adopted on each feature layer in parallel. Although the shift module incurs data movement, the shift operation is conceptually zero FLOP. Thus, adding this module increases computational cost only slightly.

\subsection{Normality/Anomaly Decision during Testing} \label{section3.3}


As only normal events are used to train the predictor without learning the decision process, a decision criterion for classifying video streams is necessary. 
The discriminating feature for the classifier is based on PSNR computed by comparing the predicted frame to the ground truth.
Our model, by design, should produce high PSNR on normal events while it would result in low PSNR on abnormal events. Thus, normality or abnormality for video streams at test time is determined by PSNR based scores with a threshold. We compute the PSNR between the ground truth and prediction of a future frame as

\begin{equation} \label{eq:5}
\text{PSNR}(\hat{I_t}, I_t)=10\log_{10}\frac{\max(\hat{I_t})}{||\hat{I}_t-I_t||^2_2}.
\end{equation}

After computing PSNR of each frame, we normalize PSNR of all the frames in each video to the range of [0, 1] by min-max normalization as a score defined by~\cite{mathieu2015deep}.

\begin{equation} \label{eq:6}
S(t)=\frac{\text{PSNR}(\hat{I_t}, I_t)-\min_t\text{PSNR}(\hat{I_t}, I_t)}{\max_t\text{PSNR}(\hat{I_t}, I_t)-\min_t\text{PSNR}(\hat{I_t}, I_t)}.
\end{equation}
Thus, normality or abnormality for a frame is determined according to the score \(S(t)\) and a threshold \(\gamma\) as

\begin{equation} \label{eq:1}
R(S(t))= \begin{cases}
    0, & \text{\(S(t)\)$<\gamma$}, \\
    1, & \text{\(S(t)\)$>\gamma$},
  \end{cases}
\end{equation}
where 0 and 1 from an abnormal events indicator \(R(\cdot)\) indicate normal and abnormal events, respectively.

\begin{table*}[!t]
\begin{center}
\begin{tabular}{p{2.5cm}|cc|c|cC{2.5cm}C{2.5cm}}
\hline
Model & Encoder & Decoder & Shift & AUC (\%) $\uparrow$ & GFLOPS $\downarrow$ & \#MParam. $\downarrow$\\
\hline
\hline
Baseline (U-Net) &  &  & &  95.80 & 44.21 & 13.29 \\
\hline \multirow{2}{2em}{\text{CPNet-0.75}}
& \checkmark &  &  & \textbf{96.55} & 33.34 (75.4\%) & 8.65 (65.1\%) \\
& \checkmark &  & \checkmark & 96.24 & 33.34 (75.4\%) & 8.65 (65.1\%) \\
\hline \multirow{2}{2em}{\text{CPNet-0.37}}
& \checkmark & \checkmark &  & 94.91 & \textbf{16.70 (37.8\%)} & \textbf{3.42 (25.7\%)} \\
& \checkmark & \checkmark & \checkmark & 96.13 & \textbf{16.70 (37.8\%)} & \textbf{3.42 (25.7\%)} \\
\hline
\end{tabular}
\end{center}
\caption{Performance (AUC) and efficiency (GFLOPs and \#MParam.) comparisons to the baseline U-Net on Ped2 dataset. Checkmarks in `Encoder' and `Decoder' indicate the application of the smaller parallel U-Net while unmarked parts keep identical to the conventional U-Net. ``Shift'' denotes the shift module.}
\label{Tab:ablation}
\end{table*}

\section{Experiment}

To evaluate the proposed method, experiments are conducted on Ped2~\cite{sabokrou2017deep} and CUHK Avenue~\cite{lu2013abnormal} datasets.  We introduce datasets and implementation details first, then we proceed to describe series of experiments including qualitative and quantitative analyses.


\subsection{Datasets}

\paragraph{The UCSD Pedestrian 2 (Ped2) Dataset} contains 16 training videos and 12 testing videos with 12 abnormal events. All of these abnormal cases are about presence of vehicles, such as bicycles, automobiles or skateboards, with their riders in a pedestrian area. The resolution of each video is 360$\times$240.

\paragraph{CUHK Avenue Dataset} contains 16 training and 21 testing videos with 47 abnormal events in total, including running and throwing of objects. The resolution of each video is 640$\times$360.

\subsection{Implementation Details}
Our method is implemented on Pytorch~\cite{pytorch}. Each video frame is resized to the size of 256$\times$256 and normalized to the range of [-1, 1]. We use the Adam optimizer~\cite{kingma2014adam} with \(\beta_1 = 0.9\) and \(\beta_2 = 0.999\). The initial learning rate is set to 2\(e\)-4 and we utilize a cosine annealing~\cite{loshchilov2016sgdr} as the learning rate scheduler. 
Our model is trained under one RTX TITAN with 4 mini-batch sizes for 60 epochs. All models are trained end-to-end and it takes about 1 and 9 hours for UCSD Ped2 and CUHK Avenue, respectively.

\subsection{Evaluation Metric}
Following the works~\cite{luo2017revisit,liu2018future}, we exploit the frame-level Area Under Curve (AUC) as the measurement for performance evaluation. It is obtained by calculating the area under the Receiver Operation Characteristic (ROC) with a varying threshold.

\subsection{Ablation Study}  \label{sec:ablation}

We conducted an ablation analysis to observe the efficiency and efficacy of our proposed CPNet. First, we evaluate different settings of CPNet by dividing the baseline U-Net into four parallel paths and applying the shift operation on the encoder part only or both of the encoder and decoder parts. 


Table~\ref{Tab:ablation} shows the results of AUC, GFLOPs and the number of parameters on the baseline and CPNet with its variants. The model CPNet-0.75~\footnote{The suffix `-0.75' means the percentage of GFLOPs to the baseline.}, which processes each frame independently on the encoder part only, results in the best AUC (96.55\%) with 75.4\% of GFLOPs and 65.1\% of the parameter number compared to the baseline. The shift module does not improve the CPNet-0.75 as its decoder part is capable of capturing temporal relationships, where a stack of encoded feature maps is processed as the baseline U-Net.

When the division is applied on both the encoder and decoder parts, the model CPNet-0.37 results 37.8\% and 25.7\% in terms of GFLOPs and the parameter number compared to the baseline. It also requires half GFLOPs rather than CPNet-0.75.
Without the shift module, CPNet-0.37 resulted in a significant performance drop as it struggles to capture temporal relationships and makes poor predictions. With the shift module, however, its performance improves to 96.13\%, which is equivalent to that of the heavier networks CPNet-0.75 and better than the baseline.

Table~\ref{Tab:score} shows the PSNR and score \(S(t)\) on the baseline and CPNet-0.37.
The baseline achieves the highest average PSNR regardless of normal and abnormal events. It highlights that the baseline prediction results in a small margin on score \(S(t)\) between normal and abnormal events.
We further observe the prediction performance on our CPNet-0.37 with and without the shift module. According to PSNR, the shift module delivers large improvements on normal events but slight improvements on abnormal events. Thus, it yields larger margins on scores \(S(t)\) so that \(S(t)\) can be an effective discriminating feature for the task.

\begin{table}[!t]
\begin{center}
\begin{tabular}{p{2cm}|C{2cm}|C{2cm}}
\hline
Model & PSNR & \(S(t)\) \\
\hline
\hline \multirow{2}{2em}{\text{Baseline}}
 & 41.20 / 38.21 & 0.74 / 0.55 \\
 & (2.99) & (0.19) \\
\hline \multirow{2}{2em}{\text{CPNet-0.37}}
 & 39.29 / 36.91 & 0.66 / 0.50 \\
 & (2.38) & (0.16) \\
\hline \multirow{2}{2em}{\text{CPNet-0.37} \text{ + Shift}}
 & 40.14 / 37.04 & 0.70 / 0.49 \\
 & (3.10) & (0.21) \\
\hline
\end{tabular}
\end{center}
\caption{PSNR and score. Left and right numbers in each cell denote the average value on normal and abnormal events, respectively. The numbers within `( )' denote their margins.}
\label{Tab:score}
\end{table}

\subsection{Comparison with the state-of-the-art methods}
\paragraph{Quantitative Results.} Results of state-of-the-art methods are summarized in Table~\ref{Tab:sota_result}. Notice that MNAD-P performs best among the methods compared but requires 51.63GFLOPs (over 3 times of CPNet-0.37) and 15.65M Parameters (over 4.5 times of CPNet-0.37) due to the extra memory networks in the design. Our model notably outperforms or is comparable to the U-Net based methods~\cite{liu2018future,dong2020dual,tang2020integrating}. As Tang~\textit{et al.}~\cite{tang2020integrating} delivers equivalent AUC on both Ped2 and Avenue datasets, it requires far more computational cost as it implements both reconstruction and prediction tasks with two U-Nets.

\begin{table}[!t]
\begin{center}
\begin{tabular}{c|C{1.5cm}|C{1.5cm}}
\hline
Methods & Ped2 & Avenue \\
\hline
\hline
MPPCA~\cite{kim2009observe} & 69.3 & - \\
MPPCA+SFA~\cite{kim2009observe} & 61.3 & - \\
MDT~\cite{mahadevan2010anomaly} & 82.9 & - \\
AMDN~\cite{xu2017detecting} & 90.8 & - \\
Unmasking~\cite{tudor2017unmasking} & 82.2 & 80.6 \\
MT-FRCN~\cite{hinami2017joint} & 92.2 & - \\
AMC~\cite{nguyen2019anomaly} & 96.2 & 86.9 \\
ConvAE~\cite{hasan2016learning} & 85.0 & 80.0 \\
TSC~\cite{luo2017revisit} & 91.0 & 80.6 \\
StackRNN~\cite{luo2017revisit} & 92.2 & 81.7 \\
AbnormalGAN~\cite{ravanbakhsh2017abnormal} & 93.5 & - \\
Frame-Pred~\cite{liu2018future} & 95.4 & 85.1\\
MemAE~\cite{gong2019memorizing} & 94.1 & 83.3 \\
Dong~\textit{et al.}~\cite{dong2020dual} & 95.6 & 84.9 \\
MNAD-R~\cite{park2020learning} & 90.2 & 82.8 \\
MNAD-P~\cite{park2020learning} & \textbf{97.0} & \textbf{88.5} \\
Tang~\textit{et al.}~\cite{tang2020integrating} & 96.2 & 85.1 \\
\hline
Baseline & 95.8 & 84.3 \\
CPNet-0.75 & \underline{96.6} & \underline{85.2} \\
CPNet-0.37 & 96.1 & 85.1 \\
\hline
\end{tabular}
\end{center}
\caption{Quantitative comparison with the state-of-the-art methods for anomaly detection. The numbers denote the average AUC (\%). The best and the second best performances are indicated in bold and underlined, respectively.}
\label{Tab:sota_result}
\end{table}

\begin{figure}[!t]
\begin{center}
    \centering 
    \includegraphics[width=0.46\textwidth]{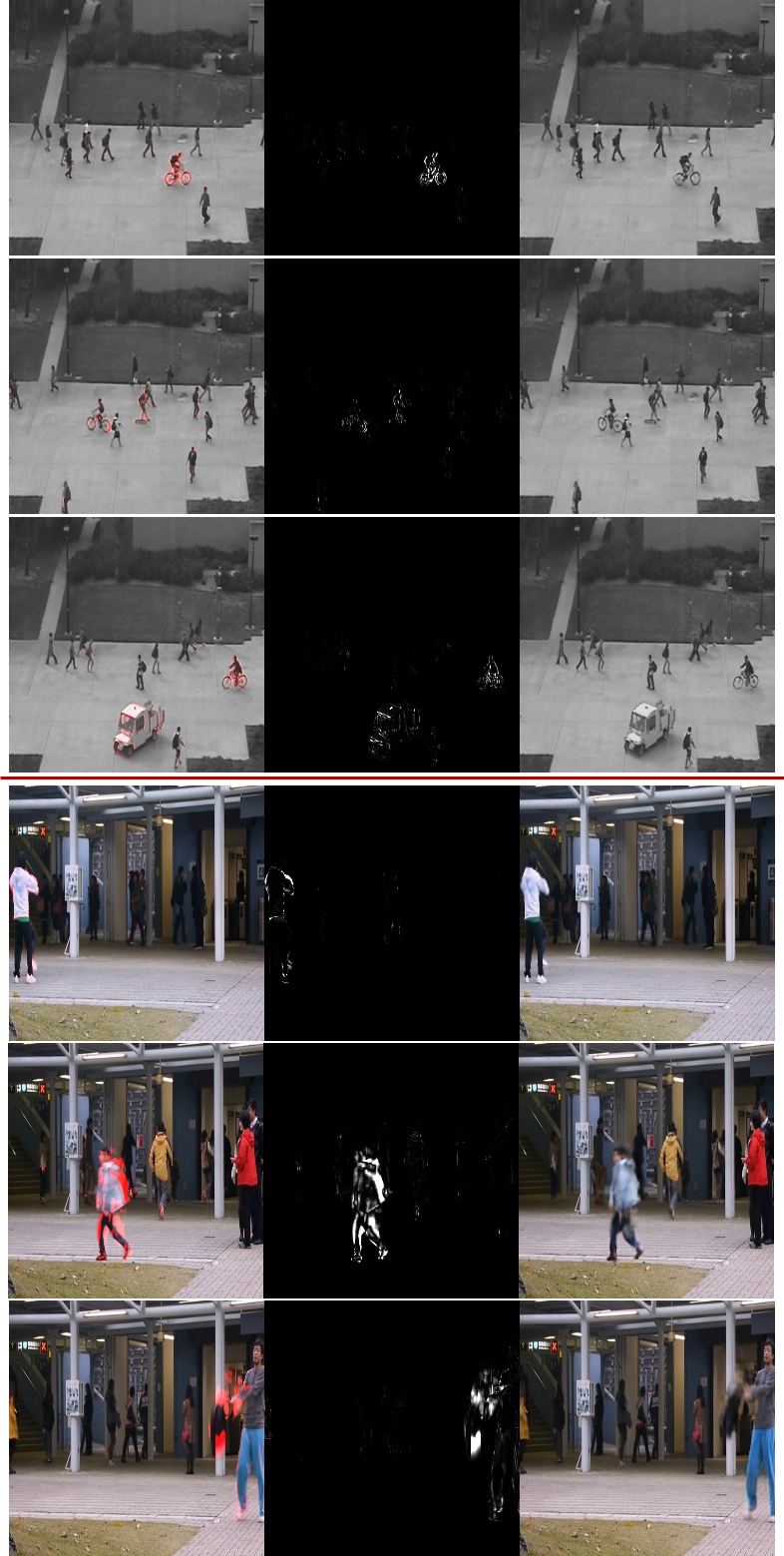}
\end{center}
\centering\caption{Qualitative results on Ped2 (top three) and Avenue (bottom three) datasets. Ground truth frames with abnormal regions are indicated in red (left). Prediction error maps (middle) and predicted future frames (right) are listed for some abnormal events, such as riding a bicycle, riding a skateboard and driving a vehicle on Ped2 dataset, and loitering, running and throwing a bag on Avenue dataset.}
\label{fig:Ped2_vis}
\end{figure}

\paragraph{Qualitative Result.} Figure~\ref{fig:Ped2_vis} shows anomaly detection results of CPNet-0.37 on Ped2 and Avenue datasets. They show the ground truth of future frames with abnormal regions denoted in red, prediction error maps and predicted future frames.

\begin{figure}[t]
\begin{center}
    \centering 
    \includegraphics[width=0.44\textwidth]{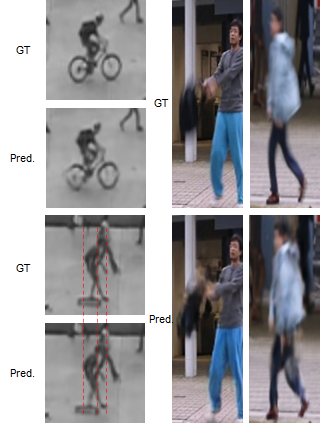}
\end{center}
\centering\caption{Illustration of abnormal regions in ground truth and predicted future frames of some examples in Figure~\ref{fig:Ped2_vis}. These regions present distortions in appearance (e.g., the wheels of bicycle and the bag) or shift (e.g., the skateboard and its rider).}
\label{fig:prediction}
\end{figure}

For the Ped2 dataset, our model well captures abnormal regions, i.e., riding a bicycle, riding a skateboard and driving a vehicle, as shown in Figure~\ref{fig:Ped2_vis}. Figure~\ref{fig:prediction} in the left column shows some abnormal regions of ground truth (GT) and predicted frames (Pred.) on the Ped2 dataset. Those parts of the image associated with abnormal events are clearly distorted in the predicted images, such as distorted bicycle wheels or horizontal shifting of the skateboard and its rider.

For the Avenue dataset, the abnormal regions, i.e., throwing a bag and running, are successfully detected. Figure~\ref{fig:prediction} in upper middle and upper right columns show abnormal regions of ground truth and the associated predicted frames are in the lower middle and the lower right columns respectively. In the predicted images, there are large distortions in the position of the bag and the appearance of people.

\paragraph{Frame Per Second.} Our method CPNet-0.37 achieves 85 fps for anomaly detection. It is far faster than other deep learning based state-of-the-art methods, e.g., 50fps for StackRNN~\cite{luo2017revisit}, 45 fps for MemAE~\cite{gong2019memorizing} and 25 fps for Frame-Pred~\cite{liu2018future} under the same setting.

\section{Conclusions}
In this paper, we proposed the Cross-Parallel Network (CPNet) in which the predictor handles each preceding frame independently with four smaller parallel U-Net and predicts a future frame. With \(\frac{1}{4}\) feature maps in each smaller U-Net, it only required 37\% GFLOPs and 25\% of the number of parameters compared with the conventional U-Net.  To mitigate the loss of cross-frame temporal relationships, the shift module was adopted.
The CPNet resulted in a significant reduction in computational cost while delivering equivalent performance on Ped2 and Avenue datasets compared with the state-of-the-art methods. According to the qualitative results, our method accurately detected abnormal regions over a variety of abnormal situations.

{\small
\bibliographystyle{ieee}
\bibliography{egbib}
}

\end{document}